\documentclass{article} 
\usepackage{nips13submit_e,times}
\usepackage{hyperref}
\usepackage{url}

\usepackage{amsmath}
\usepackage{graphicx}
\usepackage{cite}

\usepackage{fancyhdr}
\title{Abstraction in decision-makers \\ with limited information processing capabilities}

\author{
Tim Genewein\\
Max Planck Institute for Intelligent Systems\\
Max Planck Institute for Biolog. Cybernetics\\
72072 Tuebingen\\
Germany\\
\texttt{tim.genewein@tuebingen.mpg.de} \\
\And
Daniel A.~Braun\\
Max Planck Institute for Intelligent Systems\\
Max Planck Institute for Biolog. Cybernetics\\
72072 Tuebingen\\
Germany\\
\texttt{daniel.braun@tuebingen.mpg.de} \\
}

\nipsfinalcopy 

\begin{document}

\maketitle

\begin{abstract}
A distinctive property of human and animal intelligence is the ability to form abstractions by neglecting irrelevant information which allows to separate structure from noise. From an information theoretic point of view abstractions are desirable because they allow for very efficient information processing. In artificial systems abstractions are often implemented through computationally costly formations of groups or clusters. In this work we establish the relation between the free-energy framework for decision making and rate-distortion theory and demonstrate how the application of rate-distortion for decision-making leads to the emergence of abstractions. We argue that abstractions are induced due to a \emph{limit} in information processing capacity.
\end{abstract}

\section{Introduction}
Most scientific papers start with an \emph{abstract} that focuses on the main ideas of the work but leaves out many of the details. From an information theoretic point of view this allows for very efficient processing which is crucial if information processing capabilities are limited. In general, abstractions are formed by reducing the information content of an entity until it contains only information that is relevant for a particular purpose. This partial neglect of information can lead to different entities being treated as equal or, phrased differently, the separation of structure from noise. Consider the abstract concept of a ``chair'', where many aspects such as the size, color, material or particular shape are considered as noise that is irrelevant to the purpose of ``sitting down''.

The ability to form abstractions is thought of as a hallmark of intelligence, both in cognitive tasks and in basic sensorimotor behaviors \cite{tenenbaum2011grow,kemp2007learning,gershman2010learning,braun2010structure,braun2010astructure,genewein2012sensorimotor} Traditionally it is conceptualized as being computationally costly because particular entities have to be grouped together by neglecting irrelevant information. Here we argue that abstractions arise as a consequence of \emph{limited} computational capacity. The inability to distinguish different entities leads to the formation of abstractions. Note that this information processing limitation can be induced through limited computational capacity, but also through limited sample sizes or low signal-to-noise ratios. In this paper we study abstractions in the process of decision-making, where ``similar'' situations elicit the same behavior when partially ignoring the current situational context.

Following the work of \cite{simon1972theories} decision-making with limited information-processing resources has been studied extensively
in psychology, economics, political science, industrial organization, computer science and artificial intelligence research. In this paper we use a information-theoretic model of decision-making under resource constraints \cite{mckelvey1995quantal, wolpert2006information, kappen2005linear, peters2010relative, todorov2009efficient, theodorou2010generalized, rubin2012trading}. In particular, \cite{ortega2013thermodynamics,braun2011path,ortega2011information,ortega2012free} present a framework in which gain in expected utility is traded off against the adaptation cost of changing from an initial behavior to a posterior behavior. The variational problem that arises due to this trade-off has the same mathematical form as the minimization of a \emph{free energy} difference functional in thermodynamics. Here, we discuss the close connection between the thermodynamic decision-making framework \cite{ortega2013thermodynamics} and rate-distortion theory which is an information theoretic framework for lossy compression. The problem in lossy compression is essentially the problem of separating structure from noise and is thus highly related to finding abstractions \cite{tishby1999information,still2007structure,still2010optimal}.
In the context of decision-making the rate-distortion framework can be applied by conceptualizing the decision-maker as a channel from observations to actions \emph{with limited capacity}, which is known in economics as the framework of ``Rational Inattention'' \cite{sims2003implications}.

In the next section we discuss how the rate-distortion framework can be obtained for bounded-rational decision-makers that face a number of tasks. In Section~3 we demonstrate two simple applications to explore the type of abstractions that emerge from limited information processing capabilities. In Section~4 we summarize the findings and discuss the presented approach.

\section{Rate-distortion theory for decision-making}

\subsection{Bounded-rational decision-making}

In \cite{ortega2013thermodynamics}, a bounded-rational actor that initially follows a policy $p_0(x)$ changes its behavior to $q(x)$ in a way that optimally trades off the expected gain in utility against the transformation costs for adapting from $p_0(x)$ to $q(x)$. This trade-off is formalized by the following variational principle
\begin{equation}
\underset{q(x)}{\operatorname{argmax}}~ \Delta F[q] = \underset{q(x)}{\operatorname{argmax}}~\underbrace{\sum_x q(x)U(x)}_{\mathbf{E}_{q(x)}[U]} - \frac{1}{\beta} \underbrace{\sum_x q(x) \log \frac{q(x)}{p_0(x)}}_{D_{\mathrm{KL}}(q||p_0)} ,
\label{Eq:FreeEnergy}
\end{equation}
where $\beta$ is known as the \emph{inverse temperature} and $\Delta F$ is known as the difference in \emph{free energy}---negative free energy in physics---which is composed of the expected utility w.r.t. $q(x)$ and the Kullback-Leibler (KL) divergence between $q(x)$ and $p_0(x)$. $\beta$ acts as a conversion-factor between transformation cost (usually in nats or bits) and the expected utility.

The distribution $q(x)$ that maximizes the variational principle is given by
\begin{equation}
q(x) = \frac{1}{Z} p_0(x) e^{\beta U(x)} ,
\end{equation}
with the \emph{partition sum} $Z = \sum_{\xi}p_0(\xi) e^{\beta U(\xi)}$.

The influence of the transformation cost and thus the boundedness of the actor is governed by the parameter $\beta$ which determines ``how far'' the final behavior $q(x)$ can deviate from the initial behavior $p_0(x)$ measured in terms of KL-divergence. The perfectly rational actor that maximizes his utility can be recovered as the limit case $\beta \rightarrow \infty$ where transformation cost is ignored, whereas $\beta \rightarrow 0$ corresponds to an actor that has infinite transformation cost or no computational resources and, thus, sticks with his prior policy $p_0$. 

Note that in the notation shown here, $U(x)$ is conceptualized as a function over gains. In case $U(x)$ corresponds to a loss-function, the same variational principle allows to find the distribution $q(x)$ that optimally trades off minimum expected loss against transformation cost. In this case the argmin over $q(x)$ has to be taken and the sign of $\beta$ is inverted. In case $x$ is a continuous random variable, sums have to be replaced by the corresponding integrals.

\subsection{Multi-task decision-making with limited resources}
Consider an actor that is embedded into an environment and receives (potentially partial and noisy) information about the current state of the environment, that is the actor observes the value of a random variable $y$. This observation $y$ allows the actor to reduce uncertainty about the current state of the environment and adapt its behavior correspondingly. Formally this is expressed with the conditional distribution $p(x|y)$ over the action $x$. The thermodynamic framework for decision-making introduced in the previous section can straightforwardly be harnessed for describing such a bounded-rational agent that receives information $y$ by plugging in the conditional distribution $p(x|y)$ into Equation~\ref{Eq:FreeEnergy}
\begin{equation}
\underset{p(x|y)}{\operatorname{argmax}}~\mathbf{E}_{p(x|y)}[U_y(x)] - \frac{1}{\beta} D_{\mathrm{KL}}(p(x|y)||p_0(x)),
\label{Eq:FreeEnergyConditional}
\end{equation}
with the solution
\begin{equation}
p(x|y)=\frac{1}{Z}p_0(x)e^{\beta U_y(x)}.
\label{FreeEnergySolution}
\end{equation}

Notice that the utility function $U$ in general depends on the observation $y$, leading to $U(x,y)$, but to indicate the conditioning on a specific value of $y$ we write $U_y(x)$.

The initial distribution $p_0(x)$ can be interpreted as a default- or prior-behavior in the absence of an observation, thus we will refer to $p_0(x)$ as ``the prior''. The information processing cost is then given as the KL divergence between $p(x|y)$ and the prior $p(x)$ with the conversion factor $\beta$ that relates the units of transformation cost and the units of utility.

\subsection{The optimal prior}
In the free energy principle (Equation~\ref{Eq:FreeEnergyConditional}), the prior $p_0(x)$ is assumed to be given. A very interesting question is which prior distribution $p_0(x)$ maximizes the free energy difference $\Delta F$ for all observations $y$ \emph{on average}. To formalize this question, we extend the variational principle in Equation~\ref{Eq:FreeEnergyConditional} by taking the expectation over $y$ and the argmax over $p_0(x)$

\begin{equation*}
\underset{p_0(x)}{\operatorname{argmax}}~\sum_y p(y)~\left[\underset{p(x|y)}{\operatorname{argmax}}~\mathbf{E}_{p(x|y)}[U_y(x)] - \frac{1}{\beta} D_{\mathrm{KL}}(p(x|y)||p_0(x))\right].
\end{equation*}

The inner argmax-operator over $p(x|y)$ and the expectation over $y$ can be swapped because the variation is not over $p(y)$. With the KL-term expanded this leads to
\begin{equation*}
\underset{p_0(x),p(x|y)}{\operatorname{argmax}}~\sum_{x,y }p(x,y)U(x,y)- \frac{1}{\beta} \sum_y p(y) \sum_x p(x|y) \log \frac{p(x|y)}{p_0(x)}.
\end{equation*}

The solution to the argmax over $p_0(x)$ is given by $p_0(x)=\sum_y p(y)p(x|y)=p(x)$. (see 2.1.1 in \cite{tishby1999information} or \cite{csisz1984information}). Plugging in $p(x)$ for $p_0(x)$ yields the following variational principle for bounded-rational decision-making with a minimum average relative entropy prior
\begin{equation}
\underset{p(x|y)}{\operatorname{argmax}}~\underbrace{\sum_{x,y }p(x,y)U(x,y)}_{\mathbf{E}_{p(x,y)}[U]} - \frac{1}{\beta} \underbrace{\sum_y p(y) D_{\mathrm{KL}}(p(x|y)||p(x))}_{I(x;y)} ,
\label{Eq:RateDistortionVariational}
\end{equation}
where $I(x;y)$ is the \emph{mutual information} between $x$ and $y$. The variational problem can be interpreted as maximizing expected utility with an upper bound on the mutual information or in the dual point of view, as minimizing the mutual information between actions and observations with a lower bound on the expected utility. The problem in Equation~\ref{Eq:RateDistortionVariational} is equivalent to the problem formulation in rate-distortion theory (\cite{tishby1999information,cover2012elements,yeung2008information}), where $U(x,y)$ is usually conceptualized as a distortion function $d(x,y)$ which leads to a flip in the sign of $\beta$ and an argmin instead of an argmax.

The solution that extremizes the variational problem is given by the self-consistent equations (see \cite{tishby1999information})
\begin{align}
p(x|y)&=\frac{1}{Z}p(x)e^{\beta U_y(x)}, \label{Eq:FreeEnergySolutionII} \\
p(x)&=\sum_y p(y)p(x|y).
\label{Eq:OptimalPrior}
\end{align}

Note that the solution for the conditional distribution $p(x|y)$ in the rate-distortion problem (Equation~\ref{Eq:FreeEnergySolutionII}) is the same as the solution in the free energy case of the previous section (Equation~\ref{FreeEnergySolution}), except that the prior $p_0(x)$ is now defined as the marginal distribution $p_0(x)=p(x)$ (see Equation \ref{Eq:OptimalPrior}). This particular prior distribution minimizes the the average relative entropy between $p(x|y)$ and $p(x)$ which is the mutual information between actions $x$ and observations $y$.

In the limit-case $\beta \rightarrow \infty$ where transformation costs are ignored, $p(x|y)$ is equal to the perfectly rational policy for each value of $y$ \emph{independent} of any of the other policies and $p(x)$ becomes a mixture of these solutions. Note that if there is a subset of perfectly rational solutions that is shared among tasks, then only this subset will be assigned probability mass since it reduces mutual information (see Section~3.3) Importantly, high values of the mutual information term in Equation~\ref{Eq:RateDistortionVariational} will not lead to a penalization, which means that actions $x$ can be very informative about the observation $y$. The behavior of an actor with infinite computational resources will thus be very observation-specific.

In the case $\beta \rightarrow 0$ the mutual information between actions and observations is minimized to $I(x;y)=0$, leading to $p(x|y)=p(x)~\forall y$, the maximal abstraction where all $y$ elicit the same response.
The actor's behavior $p(x|y)$ becomes independent of the observation $y$ due to the lack in computational resources to change its behavior. Within this limitation the actor will, however, still emit actions that maximize the expected utility $\sum_{x,y} p(x) U(x,y)$.

For values of the rationality parameter $\beta$ in between these limit-cases, that is $0 < \beta < \infty$, the bounded-rational actor trades off \emph{observation-specific} actions that lead to a higher expected utility for particular observations at the cost of high mutual information between observations $y$ and actions $x$, against \emph{abstract} actions that yield a ``good'' expected utility for many observations and lead to a lower mutual information term.

An alternative interpretation, closer to the rate-distortion framework, is that the perceptual channel through which $y$ is transmitted to the actor has a limited \emph{capacity} given by $C=I(x;y)$. For large values of $\beta$, the transmission of $y$ is not severely influenced and the actor can choose the best action for this particular observation. For lower values of $\beta$ however, the actor becomes very uncertain about the true value of $y$ and has to choose abstract actions that are ``good'' under all observations which are compatible with the actor's belief over $y$.

\subsection{Computing the self-consistent solution}
The self-consistent solutions that maximize the variational principle in Equation~\ref{Eq:RateDistortionVariational} can be computed by starting with an initial distribution $p_0(x)$ and then iterating Equation~\ref{Eq:FreeEnergySolutionII} and Equation~\ref{Eq:OptimalPrior} in an alternating fashion. This procedure is well known in the rate-distortion framework as a Blahut-Arimoto-type algorithm \cite{yeung2008information,blahut1972computation}. The iteration is guaranteed to converge to a unique maximum (see 2.1.1 in \cite{tishby1999information} and \cite{csisz1984information,cover2012elements}. Note that $p_0(x)$ has to have the same support as $p(x)$. 

Implemented in a straightforward manner, the Blahut-Arimoto iterations can become computationally costly since the iterations involve evaluating the utility function for every action-observation-pair $(x,y)$ and computing the normalization constant $Z$. In case of continuous-valued random variables, closed-form analytic solutions exist only for special cases.

\section{Abstractions in multi-task decision-making}
\subsection{Problem formulation}
In the following we present the application of the rate-distortion framework for decision-making introduced in the previous section to multi-task decision problems.
We assume that we are given a number of tasks within the same environment and that the observations from the environment are fully informative about the current task, that is we observe the value of a discrete random variable $y$ corresponding to a unique task. Note that this assumption can easily be relaxed.

More formally we make the following assumptions: we are given a set of $N$ tasks $\tau=\{t_1,t_2,...t_N\}$ which define the set of observations $y \in \{y_1,y_2,...,y_N\}$ with $y_i=y_j$ if and only if $i=j$. Each task is defined through the utility function $U(x,y)$, where $x$ is an action. The action-space $x \in \mathcal{X}$ is the same for all tasks. We assume that the probability over tasks is known and given by $p(y)$.

The goal of the decision maker is to find task-specific distributions $p(x|y)$ that maximize the expected utility $\sum_{x,y} p(x|y) U(x,y)$ \emph{given} its computational constraints. This problem is formalized in the variational principle in Equation~\ref{Eq:RateDistortionVariational} with the self-consistent solutions in Equations~\ref{Eq:FreeEnergySolutionII},~\ref{Eq:OptimalPrior}. In this principle for bounded-rational decision-making, information processing costs arise from changing the prior-behavior $p(x)$ to the task-specific behavior $p(x|y)$ and are measured in terms of KL-cost in accordance with the thermodynamic framework for decision-making \cite{ortega2013thermodynamics}.

\subsection{Trading off abstraction against optimal action}
We designed the following two-task problem, to demonstrate the role of the rationality parameter $\beta$ that governs the trade-off between expected utility and mutual information. In both tasks, the action $x=[x_1,x_2]$ is one of four possible action-vectors (see Table~\ref{Tab:Exp1}). The utility for the first task is simply given by the value of the first component of the action vector, whereas the utility for the second task is the Manhattan distance between the two components of the action vector:
\begin{equation*}
  U(x,y)=\begin{cases}
    x_1 & \text{if}~y= y_1\\
    |x_1-x_2| & \text{if}~y=y_2
  \end{cases}.
\end{equation*}

The utilities for all actions are summarized in Table~\ref{Tab:Exp1}. The observation-variable $y \in {y_1,y_2}$ is fully informative about the task with the task probabilities $p(y_1)=p(y_2)=\frac{1}{2}$.

With this particular choice of utility functions and action-vectors, the maximum-utility action for one task has a utility of zero for the other task. However, there is a suboptimal action $x^*_{\mathrm{sub}}=[0.7,0]$ that leads to the second-best utility in both environments. The simulation results summarized in Table~\ref{Tab:Exp1} show that for a high value of the inverse temperature $\beta$ the decision-maker picks the maximum-utility action in each task with probability $1$. At a low value of $\beta$ the actor uses the same action distribution for both tasks due to its boundedness, resulting in $I(x;y)=0$. This leads to a maximal \emph{abstraction} over both tasks which is solved optimally by putting all the probability mass on the suboptimal action $x^*_{\mathrm{sub}}$. Note that the limit $\beta=1$ shown here is in general still far from the fully bounded limit $\beta \rightarrow 0$ --- in this particular example however lowering $\beta$ further has no effect.

\begin{table}
\begin{center}
  \begin{tabular}{ c || c | c || c | c | c || c | c | c }
   \multicolumn{1}{c}{ } & \multicolumn{2}{c||}{ } & \multicolumn{3}{c||}{$\beta = 100$} & \multicolumn{3}{c}{$\beta = 1$}\\ \hline
    $x$ & $U(x,y_1)$ & $U(x,y_2)$ & $p(x)$ & $p(x|y_1)$ & $p(x|y_2)$ & $p(x)$ & $p(x|y_1)$ & $p(x|y_2)$ \\ \hline
    $[0,0]$ & $0$ & $0$ & $0$ & $0$ & $0$ & $0$ & $0$ & $0$ \\
    $[0,1]$ & $0$ & $1$ & $0.5$ & $0$ & $1$ & $0$ & $0$ & $0$\\
    $[0.7,0]$ & $0.7$ & $0.7$ & $0$ & $0$ & $0$ & $1$ & $1$ & $1$ \\
    $[1,1]$  & $1$ & $0$ & $0.5$ & $1$ & $0$ & $0$ & $0$ & $0$ \\
  \end{tabular}
  \caption{Two-task decision problem. Possible actions and their utilities for both tasks are given in the first three columns of the table. The results of the Blahut-Arimoto iterations for a large value of $\beta$ are shown in the middle three columns. In this case the maximum-utility action for each task is picked with full certainty. The results for a small value of $\beta$ are shown in the last three columns. The decision maker does not have computational resources to change its behavior according to the task and thus always picks the suboptimal action that leads to a high utility in both tasks.}
  \label{Tab:Exp1}
\end{center}
\end{table}

\begin{figure}[hbtp]
  \centering
    \includegraphics[scale=0.8]{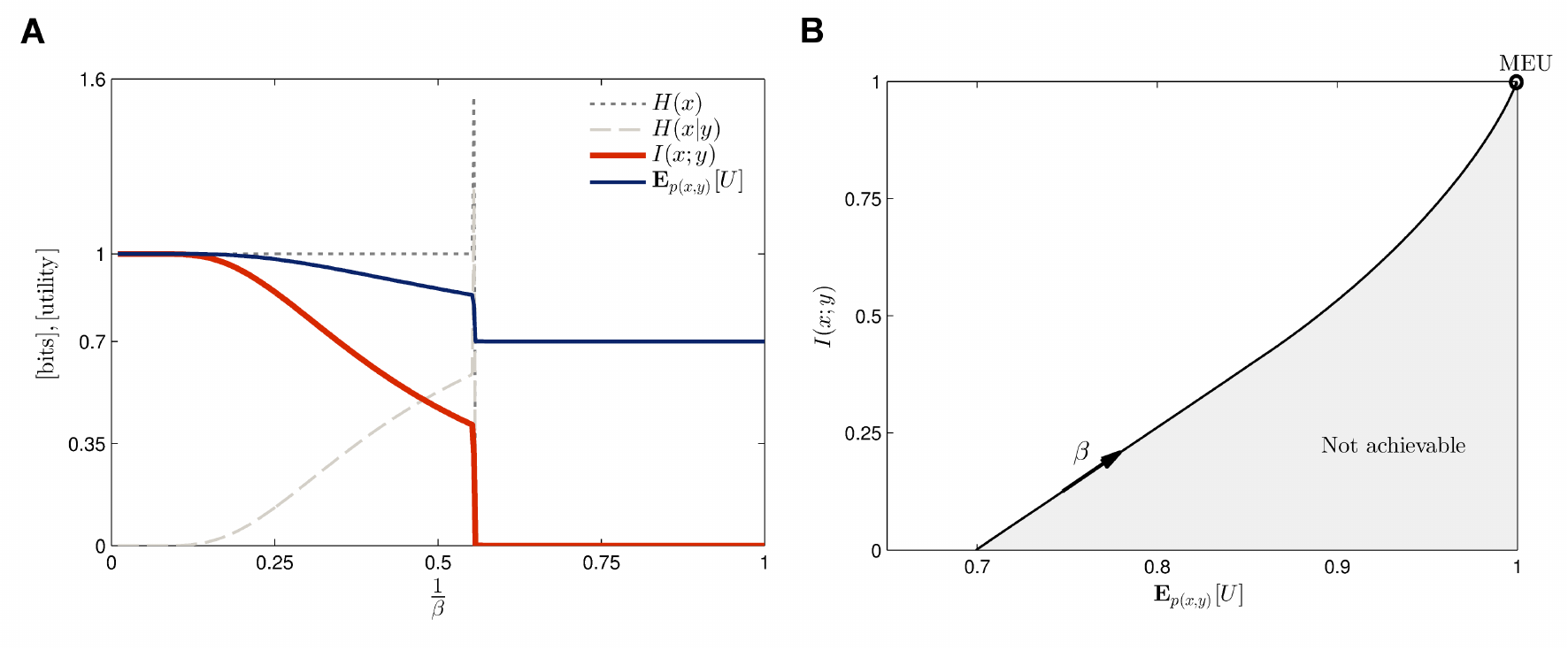}
  \caption{Transition from full rationality ($\beta \approx \infty$) to full boundedness ($\beta \approx 0$). \textbf{\textsf{A}} Trade-off between $I(x;y)=H(x)-H(x|y)$ and expected utility $\mathbf{E}_{p(x,y)}[U]$ \textbf{\textsf{B}} Rate-Utility function showing the information processing rate $I(x;y)$ as a function of the expected utility. The rate specifies the minimal average number of bits of the observation $y$ that need to be processed in order to achieve a certain expected utility. For the limit $\beta \rightarrow \infty$ the decision maker picks the maximum utility action for each environment deterministically thus following the maximum expected utility (MEU) principle.}
  \label{Fig:Exp1}
\end{figure}

Figure~\ref{Fig:Exp1} \textsf{A} shows the transition from perfect rationality to full boundedness. Starting at $\beta \approx \infty$ the entropy of the conditionals $H(x|y)$ is zero, since for a given task the actor picks the maximum-utility action with certainty. By lowering the inverse temperature $\beta$, both the mutual information $I(x;y)$ and the expected utility $\mathbf{E}_{p(x,y)}[U]$ monotonically decrease. Initially $H(x)$ stays constant, whereas $H(x|y)$ increases, which means that the actor picks the two maximum-utility actions with increasing stochasticity. At $\frac{1}{\beta} \approx 0.55$ a phase transition occurs --- the entropy $H(x)$ rapidly peaks at $1.585$bits implying that three actions are now equally probable in $p(x)$. Lowering $\beta$ further leads to a rapid drop in $H(x)$, $H(x|y)$ and $I(x;y)$ to zero bits as well as a drop in expected utility to $0.7$. The decision maker is now in the fully abstract regime, where $x^*_{\mathrm{sub}}$ is always chosen, regardless of the task.

Figure~\ref{Fig:Exp1} \textsf{B} shows the Rate-Utility function (in analogy to the rate-distortion function) where the information processing rate $I(x;y)$ is shown as a function of the expected utility. If the decision-maker is conceptualized as a communication channel between observations and actions, the rate $I(x;y)$ defines the minimal required capacity of that channel. The Rate-Utility function thus specifies the minimum required capacity for computing an action with a certain expected utility, or analogously the maximally achievable expected utility given a certain information processing capacity. Importantly, decision-makers in the shaded region are \emph{impossible}, whereas decision-makers in the white region are suboptimal with respect to their information processing capabilities.

\subsection{Changing the level of granularity}
Abstractions are formed by reducing the information content of an entity until it only contains relevant information. For a discrete random variable $x \in \mathcal{X}$ this translates into forming a partitioning over the space $\mathcal{X}$ where ``similar'' elements are grouped into the same subset of $\mathcal{X}$ and become indistinguishable within the subset. In physics changing the granularity of a partitioning to a coarser level is known as \emph{coarse-graining} which reduces the resolution of the space $\mathcal{X}$ in a nonuniform manner. In the rate-distortion framework the partitioning emerges in the shared prior $p(x)$ as a \emph{soft-partitioning} (see \cite{still2007structure}), where actions $x$ with the same average utility get the same probability mass and become essentially indistinguishable.

To demonstrate this, we use a binary grid of size $N\mathrm{x}N$, $ N=3$ where each cell of the grid can be white $x_i=0$ or colored in black $x_i=1$. Actions are particular patterns on this grid thus the actionspace becomes $x \in \left\lbrace\mathrm{binary~sequences~of~length}~N^2\right\rbrace$. The utility function defines the following three tasks: 
\begin{enumerate}
\item The utility equals the number of colored pixels, but one row and one column has to be all-white, otherwise the utility is zero. 
\item  Any pattern with exactly four colored pixels scores a utility of $+4$, all other patterns have utility zero. 
\item Any pattern with an even number of colored pixels scores a utility equal to the total number of colored pixels; all other patterns have a utility of zero.
\end{enumerate} 

\begin{figure}[hbtp]
  \centering
    \includegraphics[scale=0.8]{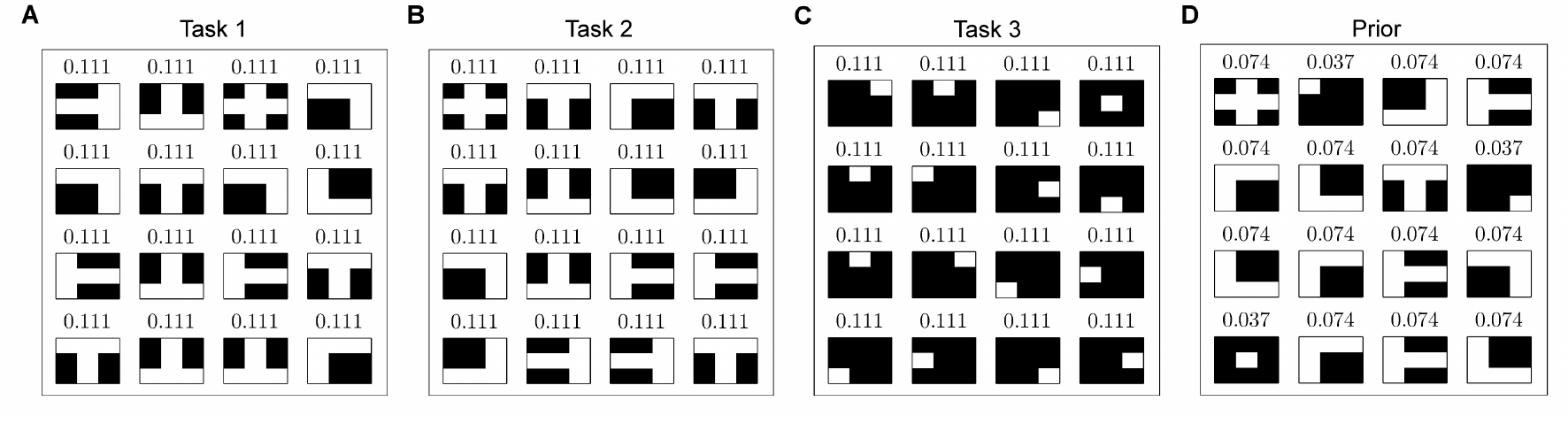}
  \caption{Sampled patterns for $\beta = 10$. The number above each pattern indicates the probability of the pattern in the corresponding distribution. \textbf{\textsf{A}} Samples for task (1) $P(x|y=1)$. All shown patterns yield maximum utility in the task. \textbf{\textsf{B}} Samples for task (2) $P(x|y=2)$. All shown patterns yield maximum utility in this task, however task (2) has more patterns that would potentially lead to maximum utility---only the subset that coincides with the maximum utility patterns in (1) has nonzero probability though. This is a consequence of sharing the same prior $p(x)$ and mutual information minimization. \textbf{\textsf{C}} Samples for task (3) $P(x|y=3)$. The patterns in task (1) and (2) would also have a nonzero probability in task (3), but the sampled patterns shown here yield twice the utility and have thus all the probability mass. \textbf{\textsf{D}} Samples from the shared prior $P(x)$. The prior is a mixture over the patterns shown in the conditional distributions.}
  \label{Fig:Exp2}
\end{figure}

Figure~\ref{Fig:Exp2} shows $16$ samples each from the conditionals $p(x|y)$ for each task and the prior $p(x)$ for $\beta = 10$. Since the inverse temperature is high, all the samples with nonzero probability are actions that yield maximum utility in their particular task. Note that the patterns that lead to a maximum utility in task (1) are a subset of the patterns that lead to maximum utility in task (2) but also lead to a nonzero utility in task (3). Since transformation costs are mostly ignored in this case, the patterns appearing for task (3) are very different from the patterns in task (1). Note however that the additional patterns in task (2) that would also lead to maximum utility are assigned a probability of zero. The subset of patterns which are also optimal in task (1) is sufficient to achieve maximum expected utility and by not including the additional ``specialized'' patterns for task (2) the mutual information can be reduced significantly. The prior $p(x)$ consists essentially of two kinds of patterns: the ones that are optimal in task (1) and (2) simultaneously and the patterns that are optimal in task (3). The first two tasks have essentially become indistinguishable because the actor will respond with exactly the same action-distribution.

\begin{figure}[hbtp]
  \centering
    \includegraphics[scale=0.8]{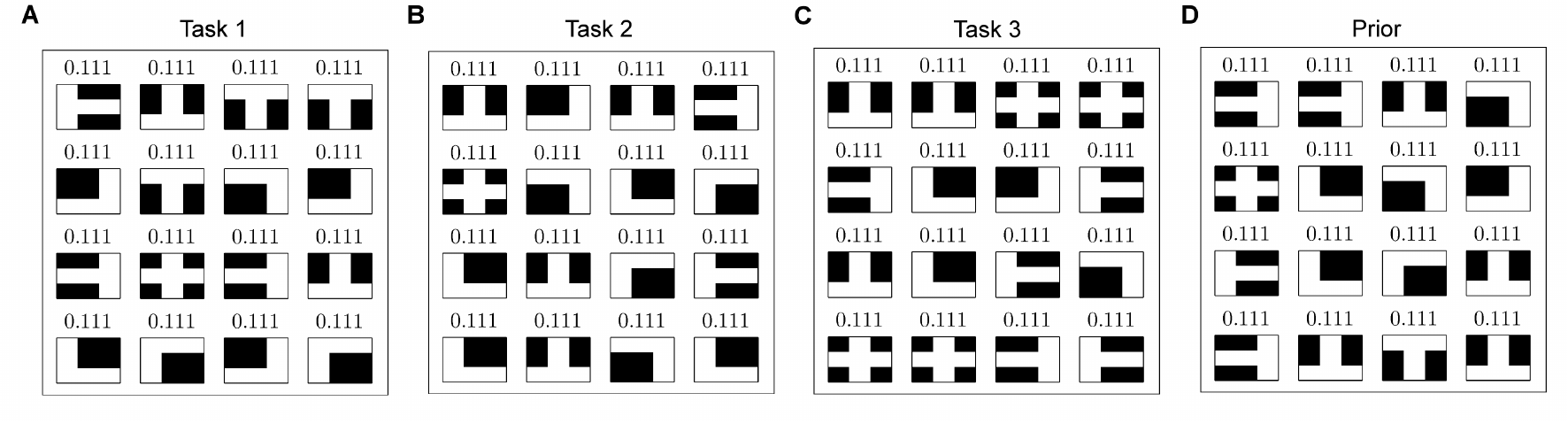}
  \caption{Sampled patterns for $\beta = 0.1$. The number above each pattern indicates the probability of the pattern in the corresponding distribution. \textbf{\textsf{A}} Samples for task (1) $P(x|y=1)$. \textbf{\textsf{B}} Samples for task (2) $P(x|y=2)$. \textbf{\textsf{C}} Samples for task (3) $P(x|y=3)$. Compared to the case $\beta = 10$ in Figure~\ref{Fig:Exp2}, the increased weight of the mutual information term $I(x;y)$ has led to the selection of suboptimal actions in task (3), similar to the previous simulation. \textbf{\textsf{D}} Samples from the shared prior $P(x)$. In the fully abstract regime all conditional distributions are exactly equal to the prior $p(x)$, leading to $I(x;y)=0$.}
  \label{Fig:Exp3}
\end{figure}

By lowering the inverse temperature to $\beta = 0.1$ (see Figure~\ref{Fig:Exp3}), the mutual information constraint gets more weight and suboptimal patterns are picked for task (3), similar to the simulation in the previous section. The behavior of the actor has now become indistinguishable for all three tasks at the expense of a lower expected utility. Importantly, the effective resolution of the prior $p(x)$ has reduced from two distinct sets of patterns to a single set of indistinguishable patterns (in terms of their expected utility). The level of granularity of the prior has been reduced even further.

\section{Discussion \& Conclusions}
In this work, we discussed the connection between the thermodynamic framework \cite{ortega2013thermodynamics} for decision-making with information processing costs and rate-distortion theory. This connection implies a novel interpretation of the rate-distortion framework for multi-task bounded-rational decision-making. Importantly, abstractions emerge naturally in this framework due to \emph{limited} information processing capabilities. The authors in \cite{van2013informational} find a very similar emergence of ``natural abstractions'' and ``ritualized behavior'' when studying goal-directed behavior in the MPD case using the \emph{Relevant Information} method, which is a particular application of rate-distortion theory. 

Although not shown here, the approach presented in this paper straightforwardly carries over to an inference case by treating $y$ as observations and $x$ as the belief-state. In the inference case, limited information processing capacities make it impossible to detect certain patterns which in turn renders different entities indistinguishable, leading to the formation of abstractions. This idea has been explored previously in \cite{still2007structure,still2010optimal}. Both, the work just mentioned and our work are inspired by the Information Bottleneck Method \cite{tishby1999information}, which is mathematically very similar to the rate-distortion problem (with a particular choice of distortion function) and thus also to the approach presented here.

Note that limited information processing capabilities can arise for various reasons. The most obvious reason, perhaps, is the lack of computational power which is in many cases equivalent to certain time-constraints (such as reaction times) or memory constraints. Other reasons for information processing limits are small sample sizes or low signal-to-noise ratios that put an upper limit on the mutual information independent of available computational power.

In the approach presented here, we assume that the decision-maker draws samples from $p(x)$. Responding to a certain task with a sample from $p(x|y)$ could then be implemented for instance with a rejection sampling procedure. The prior $p(x)$ will then be the proposal-distribution that has the highest average acceptance rate over all tasks $y$. The computational cost of finding $p(x)$ is not part of the current framework. These implications have to be explored in further work.

\subsubsection*{Acknowledgments}

This study was supported by the DFG, Emmy Noether grant BR4164/1-1.

\bibliographystyle{unsrt}
\small{
\bibliography{references}
}

\end{document}